%% file: main.tex
\documentclass[
hf,
]{ceurart}

\usepackage{listings}
\begin{document}

\copyrightyear{2024}
\copyrightclause{Copyright for this paper by its authors.
  Use permitted under Creative Commons License Attribution 4.0
  International (CC BY 4.0).}
\conference{XAI 2024 - The 2nd World Conference on eXplainable Artificial Intelligence}

\title{Patch-based Intuitive Multimodal Prototypes Network (PIMPNet) for Alzheimer's Disease classification}

\author[1,2]{Lisa Anita \mbox{De~}Santi}[%
orcid=0000-0001-7239-4270,
email=lisa.desanti@pdh.unipi.it,
]
\cormark[1]
\address[1]{Department of Information Engineering, University of Pisa, Pisa, Italy}
\address[2]{Fondazione Toscana G Monasterio - Bioengineering Unit, Pisa, Italy}
\cortext[1]{Corresponding author.}

\author[3,4]{Jörg Schlötterer}[%
orcid=0000-0002-3678-0390,
email=joerg.schloetterer@uni-marburg.de,
]
\address[3]{University of Marburg, Marburg, Germany}
\address[4]{University of Mannheim, Mannheim, Germany}

\author[5]{Meike Nauta}[%
orcid=0000-0002-0558-3810,
email=m.nauta@datacation.nl,
]
\address[5]{Datacation, Eindhoven, Netherlands}

\author[2]{Vincenzo Positano}[%
orcid=0000-0001-6955-9572,
email=positano@ftgm.it,
]

\author[3]{Christin	Seifert}[%
orcid=0000-0002-6776-3868,
email=christin.seifert@uni-marburg.de,
]

\begin{abstract}
  Volumetric neuroimaging examinations like structural Magnetic Resonance Imaging (sMRI) are routinely applied to support the clinical diagnosis of dementia like Alzheimer's Disease (AD).
  Neuroradiologists examine 3D sMRI to detect and monitor abnormalities in brain morphology due to AD, like global and/or local brain atrophy and shape alteration of characteristic structures.
  There is a strong research interest in developing diagnostic systems based on Deep Learning (DL) models to analyse sMRI for AD.
  However, anatomical information extracted from an sMRI examination needs to be interpreted together with patient's age to distinguish AD patterns from the regular alteration due to a normal ageing process.
  In this context, part-prototype neural networks integrate the computational advantages of DL in an interpretable-by-design architecture and showed promising results in medical imaging applications. 
  We present PIMPNet, the first interpretable multimodal model for 3D images and demographics applied to the binary classification of AD from 3D sMRI and patient's age.
  Despite age prototypes do not improve predictive performance compared to the single modality model, this lays the foundation for future work in the direction of the model's design and multimodal prototype training process. 
  
\end{abstract}


\maketitle

\section{Introduction}
There is a significant research interest in supporting Alzheimer's Disease (AD) diagnosis with Deep Learning (DL) models~\cite{Ebrahimighahnavieh2020}.
Existing diagnostic guidelines often integrate the clinical evaluation of the patient with structural Magnetic Resonance Imaging (sMRI), to detect pathological brain patterns like gray matter atrophy. 

Brain alterations in sMRI might support the early and differential diagnosis and the prediction of disease's progression.
There are sets of common practices used for analysing sMRI acquisition, but there are still no universally accepted methods ~\cite{DeSanti2023,Chandra2019,Vemuri2010}.
In addition, information collected from sMRI should be interpreted together with the patient's age, as there are anatomical brain changes due to the physiological ageing process~\cite{Zhao2019, Sivera2019}.

DL architectures can facilitate the analysis of neuroimaging data, and might be able to identify unconventional AD subtypes and extract yet unknown image-based biomarkers~\cite{Bohle2019,Khojaste2022}.
Prototypical-Part (PP) networks combine the advantages of DL models in an interpretable-by-design architecture, and are collecting interesting results in medical imaging applications where the black-box nature of standard DL models poses controversy~\cite{Longo2024}.

There are currently different variants of PP networks, including PIPNet~\cite{Nauta2023_PIPNet}, originally applied to 2D images and then extended to handle 3D scans~\cite{DeSanti2024}.
PIPNet showed appealing properties in the medical imaging domain~\cite{Nauta2023_PIPNet_MI}, including a reduced number of part-prototypes, semantic significance of learned prototypes, and ability to cope with Out-of-Distribution data (which might be particularly useful in dementia diagnosis, where unusual neurodegeneration pattern are reported~\cite{Vemuri2010}). 
However, sMRI data should be interpreted together with patients' demographics to discern age-related image alteration from pathological alteration, and existing PP models cannot be directly applied to perform this task.
Adding non-image prototypes to the standard PP architecture is a non-trivial task, and there are no unique strategies available. 
There are some works which integrate the concept of learning prototypes from multiple modalities which are based on the concatenation (\textit{deterministic prototypes}) or on the multimodal feature extraction (\textit{shifted prototypes}). 
However, available models cannot be applied to our task, as specifically designed for images and textual data~\cite{Ma2022}.

We present Patch-based Intuitive Multimodal Prototypes Network (PIMPNet), the first multimodal prototype classifier which learns 3D image part-prototypes and prototypical values from structured data, to predict patient's cognitive level in AD from sMRI and age values.

\section{Method} 
\begin{figure}
  \centering
  \includegraphics[width=\linewidth]{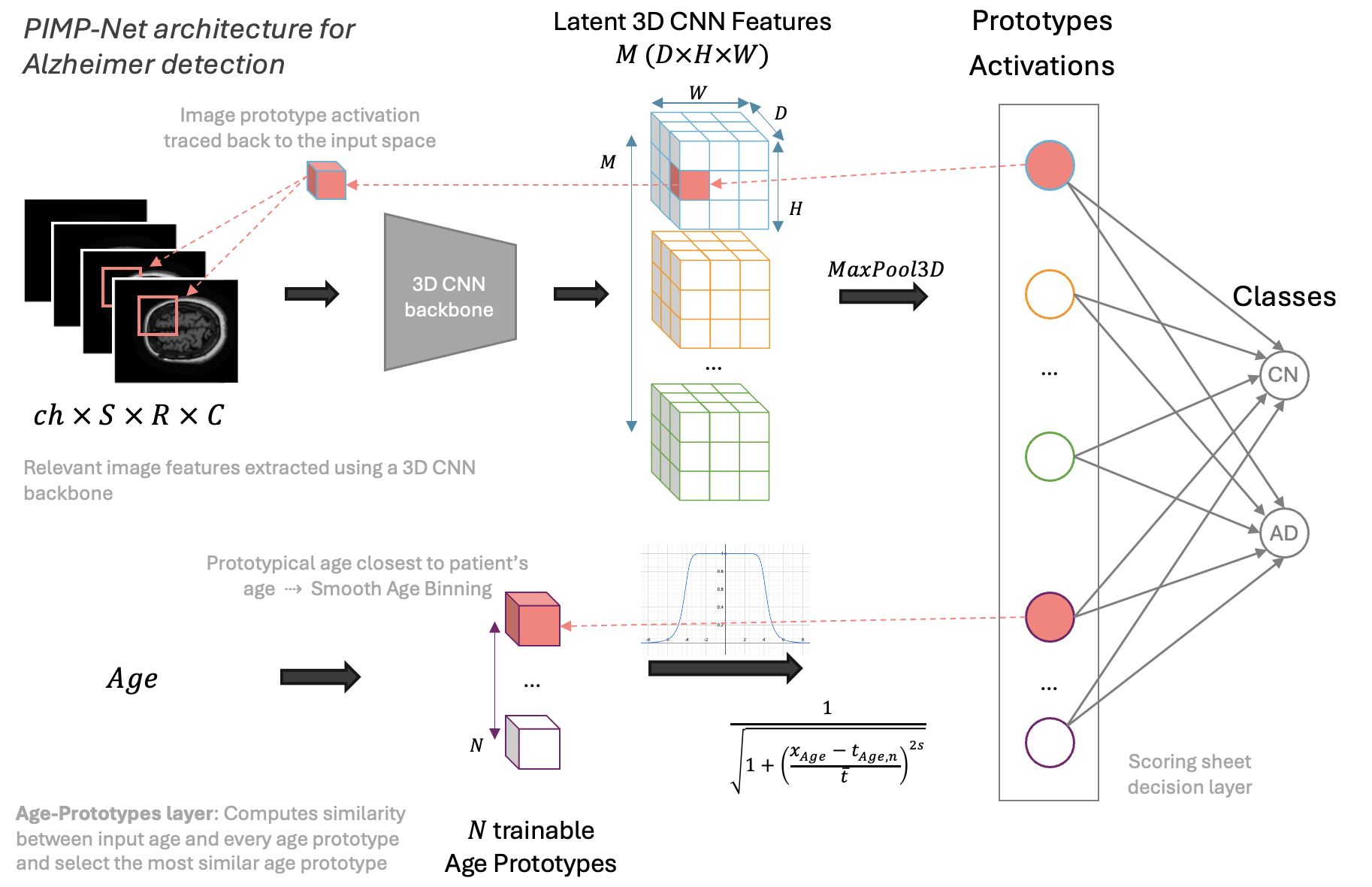}
  \caption{PIMPNet architecture}
  \label{fig:mmpipnet}
\end{figure}

This section introduces the architecture (cf. Sect.~\ref{sec:mmpipnet_architecture} and Fig.~\ref{fig:mmpipnet}) and the training process (Sect.~\ref{sec:mmpipnet_training}), of PIMPNet.

\subsection{Proposed Model: PIMPNet} 
\label{sec:mmpipnet_architecture}
We propose an \textit{age-prototypes layer} integrated into the original PIPNet 3D model~\cite{DeSanti2024} to create our multimodal architecture.
In contrast to ``ordinary'' age-binning for the inclusion of age information, the \textit{age-prototypes layer} has the advantages of: (i) being able to learn important age values for the diagnostic task performed (which might not be equally distributed, and might not be easily identifiable in priors); (ii) not to assign different age bins to two patients of similar age close to the bin boundary.

Our PIMPNet has an \textbf{input layer} which takes the 3D image $\mathbf{x}_{img}\in\mathbb{R}^{ch\times S\times R\times C}$ and the age $\mathbf{x}_{age}\in\mathbb{R}^1$ as input, where $ch,S,R,C$ respectively represents the number of channels, slices, rows and columns of the input image volume.
Image $\mathbf{x}_{img}$ and age $\mathbf{x}_{age}$ are processed in parallel. 
A \textbf{CNN backbone} processes $\mathbf{x}_{img}$, $\mathbf{z}=f(\mathbf{x}_{img};\mathbf{w}_f)$ extracting $M$ 3-dimensional $(D\times H\times W)$ feature maps, where $\mathbf{z}_{m,d,h,w}$ represents the activation of image-prototype $m$ at patch location $d,h,w$. 
Next, a \textbf{3D max-pooling} applied to every feature map extracts $M$ image prototype presence scores $\mathbf{p}_{img} \in [0., 1.]^M$, where $\mathbf{p}_{img,m}$ measures the presence of the image prototype $m$ in the input image. This defines the \textbf{image-prototypes layer}.

In parallel, we have the \textbf{age-prototypes layer}, constituted by $N$ trainable 1-dimensional tensors $\mathbf{t}_{age,n}\in\mathbb{R}^{1\times N}$, which aims to learn prototypical age values for the classification task. 
This layer computes age prototype presence scores $\mathbf{p}_{age} \in [0., 1.]^N$, a similarity measurement between the input age to every age prototype, defining a smooth age binning \footnote{The similarity function employed is inspired by the magnitude of a Butterworth filter~\cite{butterworth1930}. In preliminary experiments, we used an exponential similarity function as in ProtoTree: $\mathbf{p}_{age,n} = exp(-||\mathbf{x}_{age}-\mathbf{t}_{age,n}||)$, but as $exp(-2)\approx0.13$, a $2$ years age difference would already result in little similarity, which is not in line with domain knowledge about age relevance for Alzheimer's disease.}:
\begin{equation}
    \mathbf{p}_{age,n} = \frac{1}{\sqrt{1+\left(\frac{\mathbf{x}_{age}-\mathbf{t}_{age,n}}{\overline{t}}\right )^{2s}}},  
\end{equation} where $\mathbf{t}_{age}$ are trainable parameters and  $\overline{t}$ and $s$ are hyper-parameters which regulate the band and the slope of the similarity function.
We then have a \textbf{prototypes layer} which concatenates the image and age prototype presence scores obtaining a layer of $L=M+N$ prototypes $\mathbf{p} \in [0., 1.]^L$ $\mathbf{p}=concat(\mathbf{p}_{img},\mathbf{p}_{age})$.
The final classification is performed by a \textbf{sparse linear positive layer} $\mathbf{w_c}\in\mathbb{R}^{L\times K}_{\geq 0}$ which connects image and age prototypes to the $K$ classes acting as a scoring sheet system. 
The $K$ \textbf{class output scores} are given by the sum of the prototypes' presence score weighted for the contribution of prototype $l$ to class $k$ $\mathbf{w_c}^{l,k}$, i.e., $\mathbf{o}=\mathbf{p}\mathbf{w_c}$, where $\mathbf{o}$ is $1\times K$ and $\mathbf{o}_k=\sum_{l=1}^L \mathbf{p}_l\mathbf{w_c}^{l,k}$.
PIMPNet returns the output class only using the most activated age-prototype, i.e., closest to the patient's age according to the similarity metric\footnote{We selected only the most activated age prototype during inference (not during the optimization process).}.

\subsection{PIMPNet Training}
\label{sec:mmpipnet_training}
We optimize PIMPNet's parameters by integrating the training of age prototypes into the original PIPNet training process~\cite{Nauta2023_PIPNet}.
This includes two main stages: (1) Self-Supervised pre-training of Image-Prototypes, and (2) PIMPNet training.

As the original PIPNet~\cite{Nauta2023_PIPNet}, the 1st stage generates positive pairs $\mathbf{x}_{img}', \mathbf{x}_{img}''$ by applying data augmentation transformation to $\mathbf{x}_{img}$ selected so that humans consider the two views similar.
These are used to minimize the loss function $\lambda_A\mathcal{L}_A+\lambda_T\mathcal{L}_T$ by updating $\mathbf{w}_f$, where $\mathcal{L}_A=-\frac{1}{DHW}\sum\limits_{(d,h,w)\in D\times H \times W} log(\mathbf{z}'_{:,d,h,w}\cdot \mathbf{z}''_{:,d,h,w})$ is an \textit{Alignment Loss} 
which optimizes positive pairs to activate the same prototype. Together with a softmax over $\mathbf{z}_{:,d,h,w}$, the alignment results in near-binary encodings where an image patch corresponds to exactly one prototype.
$\mathcal{L}_T=-\frac{1}{M}\sum log(tanh (\sum \mathbf{p}_{img,b})+\epsilon)$ is a \textit{Tanh-Loss} used to prevent the trivial solution that one prototype node is activated on all image patches in each image in the dataset and instead activates multiple distinct prototypes per batch $b$. Only during training, output scores are calculated as $\mathbf{o}=\log((\mathbf{p}\mathbf{w}_c)^2+1)$, acting as regularization for sparsity.

The 2nd training stage includes the training of age prototypes, optimization of classification performance and image-prototypes fine-tuning for the downstream classification task.
The optimization minimizes $\lambda_A\mathcal{L}_A+\lambda_T\mathcal{L}_T+\lambda_C\mathcal{L}_C$ by updating $\mathbf{w}_f,\mathbf{t}_{age},\mathbf{w}_c$, where $\mathcal{L}_C$ is the Log-likelihood classification loss

\section{Evaluation}
We used the multimodal dataset from the Alzheimer's Disease Neuroimaging Initiative (ADNI) database\footnote{\url{https://adni.loni.usc.edu}}. 
We selected \textit{``ADNI1 Standardized Screening Data Collection for 1.5T scans''} processed with Gradwarp, B1 non-uniformity, and N3 correction, obtaining 307 CN and 243 AD sMRI brain scans and the corresponding patients' age.
We report statistics on patients' demographic of the selected ADNI cohort in Table~\ref{tab:demographic}.
\input{tables/demographics} 
We preprocessed sMRI data inspired by the pre-processing pipeline applied in previous works~\cite{Mulyadi2023}. 
We tranformed all the images to the common ICBM152 Non-Linear Symmetric 2009c standard space~\cite{Fonov2009} with affine registration. 
We selected the grey matter structures by applying the ICBM152 Non-Linear Symmetric 2009c brain mask and kept a margin of 3 from its first and last non-empty slices. 
We applied an image downsampling of 2 and we scaled all image intensities to the range [0-1] with a min-max normalization.
We implemented PIMPNet using PyTorch and MONAI\footnote{{\url{https://monai.io}}}, training our models on an Intel Core i7 5.1 MHz PC, 32 GB RAM, equipped with an NVIDIA RTX3090 GPU with 24 GB of embedded RAM.
As CNN backbones we used ResNet-18 3D pretrained on Kinetics400~\cite{Tran2017} and ConvNeXt-Tiny 3D pretrained on the  STOIC medical dataset (Study of Thoracic CT in COVID-19)~\cite{Kienzle2022}.
We finetuned PIMPNet with Adam optimizer using the same hyperparameter settings of the original PIPNet~\cite{Nauta2023_PIPNet}. 
We only reduced the batch size to $12$ to adapt it to our computational capabilities and we set the learning rate of age prototypes to $0.1$\footnote{Using the same learning rate of the original PIPNet used to train the image-prototypes (0.05) results in irrelevant updates of the Age Prototypes}.
We arbitrarily set the number of age prototypes $=5$ evenly spaced between $40$ and $90$ to cover the patients' age range of our dataset. For the age similarity function, we respectively set $\overline{t}=4$ and $s=8$\footnote{We leave an extensive hyperparameter search for learning the age prototypes for future work}.
We performed 5-fold cross-validation with patient-wise splits. $20\%$ of training images are used for validation.

We evaluated the models in terms of classification performance and with functionally grounded metrics of explainability.
Results are reported in Tables~\ref{tab:performances} and~\ref{tab:functionally_eval}.
\begin{table*}
\caption{Performances comparison between PIPNet trained on 3D sMRI and PIMPNet trained on 3D sMRI + Age averaged over 5 folds. }
\label{tab:performances}
\centering
\begin{tabular}{lccccc}
\toprule 
\textbf{Model} & \textbf{Acc} & \textbf{Bal Acc} & \textbf{SENS} & \textbf{SPEC} & \textbf{F1} \\ 
\midrule
\textit{\textbf{PIPNet}} & & & & & \\
\textit{ResNet-18 3D} & 83 $\pm$ 04 & 83 $\pm$ 04 & 86 $\pm$ 06 & 79 $\pm$ 07 & 81 $\pm$ 05 \\ 
\textit{ConvNeXt-Tiny 3D} & 65 $\pm$ 12 & 66 $\pm$ 09 & 56 $\pm$ 32 & 76 $\pm$ 15 & 66 $\pm$ 05 \\
\textit{\textbf{PIMPNet}} & & & & & \\
\textit{ResNet-18 3D} & 84 $\pm$ 04 & 83 $\pm$ 04 & 89 $\pm$ 03 & 77 $\pm$ 08 & 81 $\pm$ 05 \\ 
\textit{ConvNeXt-Tiny 3D} & 72 $\pm$ 04 & 70 $\pm$ 04 & 86 $\pm$ 10 & 55 $\pm$ 14 & 63 $\pm$ 09 \\ 
\bottomrule
\end{tabular}
\end{table*}
We compared PIMPNet performance (sMRI + age) with PIPNet-3D (sMRI only)~\cite{DeSanti2024}, to evaluate if including age information improves diagnostic performance. 
We measured performance using Accuracy (Acc), Balanced Accuracy (Bal Acc), Sensitivity (SENS, Acc of Cognitive Normal class), Specificity (SPEC, Acc of Alzheimer's Disease class), and F1 score (F1). 
We measured the Global size (GS) of the model as the total number of prototypes, and the Local size (LS) of explanations as the number of detected prototypes in a single 3D sMRI, averaged over all the images in the test set. Additionally, we report the Sparsity (Sp) of the decision layer as the percentage of zero-weights in the linear classification layer~\cite{Nauta2023_PIPNet} to assess the compactness of the prototypes-classes layer. 
We further assessed whether prototypes are consistently located in the same brain region, and the purity of the prototypes in terms of the anatomical regions included based on the CerabrA atlas annotation~\cite{Manera2020}. 
More specifically, the Prototypes Localization Consistency (LC$_p$) evaluates the differences in the coordinate centre of the prototypical part in the input image, while the Prototype Brain Entropy (H$_p$) as a measure of purity computes the Shannon Entropy of the brain regions included in the prototypical part~\cite{DeSanti2024}.

We show the learned age prototypes \textbf{$\mathbf{t}_{age}$} from five different folds (denoted as Mx where x indicates the current fold) in Table~\ref{tab:age_proto}.

\begin{table*}
\caption{Functionally-grounded evaluation of PIPNet trained on 3D sMRI and PIMPNet trained on 3D sMRI + Age averaged over 5 folds. $\uparrow$ and $\downarrow$: tendency for better values.}
\label{tab:functionally_eval}
\centering
\begin{tabular}{lccccc}
\toprule 
\textbf{Model} & \textbf{GS $\downarrow$} & \textbf{LS $\downarrow$} & \textbf{Sp $\uparrow$} & \textbf{LC$_p$ $\downarrow$} & \textbf{H$_p$ $\downarrow$} \\ 
\midrule
\textit{\textbf{ResNet-18 3D}} & & & & & \\
\textit{PIPNet} & 149 $\pm$ 18 & 73 $\pm$ 10 & 0.855 $\pm$ 0.018 & 0.008 $\pm$ 0.006 & 2.474 $\pm$ 0.249 \\ 
\textit{PIMPNet} & 143 $\pm$ 35 & 74 $\pm$ 20 & 0.861 $\pm$ 0.033 & 0.006 $\pm$ 0.006 & 2.424 $\pm$ 0.162 \\ 
\textit{\textbf{ConvNeXt-Tiny 3D}} & & & & & \\
\textit{PIPNet} & 4 $\pm$ 2 & 2 $\pm$ 1 & 0.997 $\pm$ 0.001 & 0.000 $\pm$ 0.000 & 1.803 $\pm$ 0.999 \\
\textit{PIMPNet} & 10 $\pm$ 9 & 4 $\pm$ 4 & 0.993 $\pm$ 0.002 & 0.000 $\pm$ 0.000 & 1.543 $\pm$ 0.626 \\ 
\bottomrule
\end{tabular}
\end{table*}

\input{tables/age-proto-short} 

\section{Discussion and Conclusion}
Both PIPNet and PIMPNet with the ResNet-18 3D backbone achieve higher classification performance than with the ConvNext-Tiny backbone.
Our preliminary results also show that the proposed Age-Prototype layer can learn prototypical age values; however, these do not improve classification performances compared to the baseline model. 
Our functionally-grounded evaluation of prototypes shows that all models learn prototypes consistently located in the same anatomical brain regions (low LC$_p$ values).
We also observe that the models trained with the ConvNeXt-Tiny 3D backbone have higher compactness. This might partially explain the lower performance scores (the number of prototypes learned is not enough for performing the diagnosis), but is an interesting observation for future research as such a highly compact model can be considered more interpretable than larger ones and can be easily evaluated by domain experts.
We also observe that the image prototypes of the ConvNeXt-Tiny 3D backbone are generally purer\footnote{Purity is measured w.r.t. to the annotation provided by the CerebrA atlas}. 
Despite purity being a desirable property for prototypes~\cite{Nauta2023_XAI}, because of the design of the purity metric, also a prototype which only includes the background, i.e., a clinically irrelevant prototype, will have high purity\footnote{Posterior quantitative evaluation performed w.r.t. the CerebrA atlas revealed that the test set image prototypes (averaged over the 5-folds) obtained with the ConvNeXt-Tiny backbone have a higher percentage of background voxels included compared to the ones obtained with ResNet-18 (76.6$\%$ vs 59.2$\%$)}.

In summary, we proposed PIMPNet, an interpretable multimodal prototype-based classifier.
The proposed architecture is the first prototypes-based network which performs an interpretable classification based on the detection of prototypes learned from different data modalities (3D images and age information).
We applied PIMPNet to the binary classification of Alzheimer's Disease from 3D sMRI images together with the patient's age. 
Despite the usage of age prototypes do not improve predictive performance compared to the model trained with only images, we identified different potential reasons which define the future directions of our work.
First, as the original PIPNet training paradigm includes a pretraining stage~\cite{Nauta2023_PIPNet} of image prototypes, we plan to include an age prototypes pretraining step w.r.t. the log-likelihood classification loss. 
Second, we also plan to work on the model's design. As the simple concatenation of the prototype presence score might not be able to properly represent the relationship between age and image prototypes for the downstream task, we plan to combine image and age prototypes using a different (but still interpretable) classifier than a scoring-sheet system.

\begin{acknowledgments}
  Data used in the preparation of this article was obtained from the Alzheimer’s Disease Neuroimaging Initiative (ADNI) database. The ADNI was launched in 2003 as a public-private partnership with the primary goal to test whether serial magnetic resonance imaging (MRI), positron emission tomography (PET), other biological markers, and clinical and neuropsychological assessment can be combined to measure the progression of mild cognitive impairment (MCI) and early Alzheimer's disease (AD).  
\end{acknowledgments}

\bibliography{bibliography}

\end{document}

%% file: tables/demographics.tex
\begin{table*}[h]
\caption{Patients' demographics of the selected ADNI cohort, further divided according to the clinical labels.}
\label{tab:demographic}
\centering
\begin{tabular}{lccc}
\toprule 
\textbf{Class} & N° subjects & Mean $\pm$ SD Age & Age Range \\ 
\midrule
\textit{Both} & 550 & 76 $\pm$ 6 & 55-91 \\ 
\textit{~CN} & 307 & 76 $\pm$ 5 & 60-90 \\ 
\textit{~AD} & 243 & 75 $\pm$ 8 & 55-91 \\

\bottomrule
\end{tabular}
\end{table*}

%% file: tables/age-proto-short.tex
\begin{table*}
\caption{Prototypical Age Values $\mathbf{t}_{Age,i}$ learned for folds M1, ..., M5 trained with different backbones.}
\label{tab:age_proto}
\centering
\begin{tabular}{lcccccccccc}
\toprule 
\textbf{Fold} & $\mathbf{t}_{Age,1}$ & $\mathbf{t}_{Age,2}$ & $\mathbf{t}_{Age,3}$ & $\mathbf{t}_{Age,4}$ & $\mathbf{t}_{Age,5}$ & $\mathbf{t}_{Age,1}$ & $\mathbf{t}_{Age,2}$ & $\mathbf{t}_{Age,3}$ & $\mathbf{t}_{Age,4}$ & $\mathbf{t}_{Age,5}$  \\ 
\midrule
& \multicolumn{5}{c}{\textit{ResNet-18 3D}} &  \multicolumn{5}{c}{\textit{ConvNeXt-Tiny 3D}}  \\\cmidrule(r){2-6} \cmidrule(l){7-11}
M1 & 65.77 & 65.81 & 66.14 & 76.81 & 80.99  &  56.81 & 65.00 & 64.96 & 74.13 & 85.80 \\ 
M2 & 68.46 & 69.40 & 70.38 & 77.04 & 82.38  &  55.75 & 58.39 & 64.96 & 74.32 & 85.59 \\
M3 & 66.37 & 67.27 & 67.91 & 75.87 & 81.96  &  54.86 & 56.63 & 65.21 & 74.40 & 85.11 \\
M4 & 66.72 & 66.72 & 67.07 & 77.07 & 79.75  &  58.22 & 58.59 & 66.50 & 75.88 & 89.09 \\
M5 & 66.51 & 66.52 & 67.23 & 77.37 & 80.00  &  57.79 & 66.94 & 65.44 & 72.55 & 84.58 \\
\bottomrule
\end{tabular}
\end{table*}